\documentclass[conference]{IEEEtran}
\usepackage[margin=0.75in]{geometry}

\IEEEoverridecommandlockouts

\usepackage{cite}
\usepackage{amsmath,amssymb,amsfonts}
\usepackage{algorithmic}
\usepackage{graphicx}
\usepackage{textcomp}
\usepackage{xcolor}
\usepackage{multirow}
\usepackage{float}
\usepackage{siunitx}

\newcommand{\revcomment}[2]{\textcolor{red}{#2}$^{\##1}$} 

\renewcommand{\revcomment}[2]{#2}   

\usepackage[hidelinks,
  pdftitle={A High-Payload Wall-Climbing Robot Using Passive Bistable Suction Cups},
  pdfauthor={Andrew Nguyen; Mingyuan Li; Daniel Bruder}]{hyperref}

\def\BibTeX{{\rm B\kern-.05em{\sc i\kern-.025em b}\kern-.08em
    T\kern-.1667em\lower.7ex\hbox{E}\kern-.125emX}}
\begin{document}

\title{A High-Payload Wall-Climbing Robot Using\\ Passive Bistable Suction Cups}

\author{Andrew Nguyen, Mingyuan Li, Daniel Bruder, \emph{Member, IEEE} %
\thanks{The authors are with the Mechanical Engineering Department at the University of Michigan, Ann Arbor, MI 48109, USA \{\tt\small andyng, richieli, bruderd\}@umich.edu}%
}


\maketitle

\begin{abstract}
%
Wall-climbing robots capable of scaling vertical surfaces could help automate hazardous or labor intensive tasks such as window washing, inspection, maintenance, and construction. Active adhesion methods achieve higher payload capacities, but require power to maintain their grip. Passive adhesion devices such as suction cups are an attractive option for such robots because they do not require power to maintain their grip, but they are limited by their payload capacity. This work presents a novel high-payload wall-climbing robot that utilizes passive bistable suction cups to generate adhesion without needing to be pushed into the wall. The robot features a track-based system that automatically engages and disengages bistable suction cups to achieve locomotion on smooth surfaces. The robot is able to achieve vertical wall climbing on glass, wood, metal, and painted surfaces, sideways and upside-down climbing, and is able to tow a payload of 7.940 kg (with a payload-to-weight ratio of 2.25). Project Website: \url{https://bistable-suction-cup-wall-climbing.github.io/}

\end{abstract}


\section{Introduction}
    
    Wall-climbing robots, which are capable of scaling vertical surfaces, could help automate many hazardous or labor intensive real-world tasks such as window washing \cite{miyake2006development,lee2019three,vega2019design,mir2018portable, sun2007climbing}, building and infrastructure inspection \cite{ma2024exploration}, structure maintenance \cite{yan1999development}, construction \cite{chu2010survey}, and bathroom cleaning.
    Existing wall-climbing robots, however, have limitations that reduce their effectiveness in real-world applications. Namely, they have low payload capacity due to the force limitations of their adhesion mechanisms and/or low energy efficiency due to the power requirements of their adhesion systems.
    This work presents a novel wall-climbing robot that exploits the bistable behavior of custom designed passive suction cups to generate large adhesion forces without requiring large application forces, achieving greater efficiency and payload capacity than existing wall-climbing robots.
    
    There are many different types of wall-climbing robots and previous research has divided them into categories based on the method of climbing \cite{nansai2016survey, fang2022design, fang2023advances}. 
    The methods of adhesion can be divided into two distinct categories: active adhesion and passive adhesion.

    Active adhesion means energy is required to keep a robot attached to a surface. One such method is electro-adhesion, which utilizes electric fields to generate attractive forces between a charged surface and the wall \cite{gu2018wall-climbing}. Another technique involves an Electric Ducted Fan (EDF), which creates suction in the low-pressure zone left behind the fan \cite{papadimitriou2019development}. A well-researched approach is vacuum suction, which provides a constant adhesive force \cite{lee2016series,shang2007design,guan2012modular,miyake2009vacuum}. However, the primary drawback of active adhesion is the continuous energy consumption required to maintain adhesion, which limits the operating time and design variability of robots that use this method.

    In contrast, passive adhesion methods require no energy to maintain surface adhesion. One example is the use of permanent magnets \cite{shen2005permanent}.
    Another passive approach mimics gecko-inspired directional adhesives \cite{kalouche2014inchworm,kim2008directional}, and another utilizes adhesive tape as a similar alternative \cite{murphy2007waalbot}.
    Bio-inspired methods, such as utilizing spikes to penetrate or claws to catch on surface asperities, also offer passive adhesion \cite{spenko2008biologically}.
    For smooth surfaces, suction cups offer an attractive passive adhesion method due to their low-cost and simplicity \cite{yoshida2010design,yoshida2011wall,Kawasaki2014DevelopmentOA}.
    They provide temporary adhesion to smooth surfaces without leaving residue or causing damage, making them ideal for applications requiring frequent repositioning.
    Despite their energy efficiency, however, passive methods often have limited payload capacities. For instance, the robot in \cite{yoshida2011wall} could only carry its own mass of \qty{0.5}{kg}, even though it utilized 14 redundant suction cups.
    
    \begin{figure}[H]
    \centering
    \includegraphics[width=\linewidth]{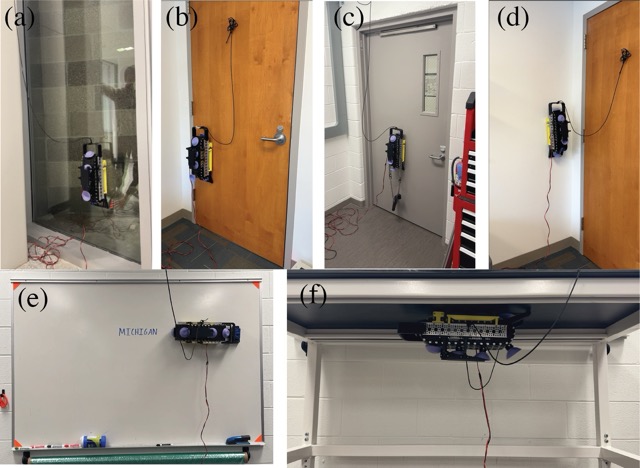} 
    \caption{
    The proposed wall climbing robot uses bistable suction cups to adhere to a variety of surfaces. It has been shown capable of (a) climbing vertically on a glass window, (b) climbing vertically on a wooden door, (c) climbing vertically on a steel door while carrying a payload, (d) climbing vertically on a painted wall, (e) wiping a whiteboard while traversing sideways on it, and (f) translating on the underside of a lab bench.}
    \label{fig:collage}
    \end{figure}

    \revcomment{}{Previous work has examined the payload capacity (and payload-to-weight ratio) of different wall-climbing mechanisms based on their adhesion method \cite{rajendran2023}. Magnetic based mechanisms have produced the highest payload-to-weight ratio of 10 \cite{Yanqiong2008902}, but they are only functional on Ferromagnetic surfaces. Active-vacuum-suction-cup based mechanisms have achieved a payload-to-weight ratio of 4.2 \cite{Sadegh07}, however they require energy to maintain adhesion.}

    This work presents a suction-cup-based method of passive adhesion that enables larger payload capacities than previous wall-climbing robots. 
    %
    A suction cup creates an ``adhesion force'' through a pressure difference between the air trapped under the cup and the surrounding air.
    Traditionally, to generate this pressure difference, a force must be applied to deform the suction cup and force air out from underneath it. 
    After this ``application force'' is removed, the air trapped beneath the suction cup has a lower density than the air outside, resulting in a lower pressure underneath the cup.
    The adhesion force this generates is directly correlated with the amount of air forced out from beneath the suction cup by the application force.
    Adhesion will be maintained until the seal between the cup and surface is broken and the pressures equalize.

    This relationship between application force and adhesion force presents a challenge for wall-climbing robots that use suction cups.
    If a robot attempts to push one of its suction cups too hard against the wall, it could overcome the adhesion force of its other suction cups and cause it to fall.
    In other words, the robot could ``push itself off of the wall''.
    If instead, the robot pushes its suction cups against the wall softly, the resulting adhesion force will be small, limiting the overall strength of adhesion and payload capacity of the robot.
    We refer to this trade-off as the ``suction cup force trade-off".

    %
    The wall-climbing robot developed in \cite{yoshida2011wall} illustrates this trade-off.
    This design utilizes a pantograph mechanism to push against the surface, causing the body of the robot to pivot around the rotating shaft and push each suction cup onto the wall.
    The pantograph mechanism cannot push against the wall too hard without causing the cups already in contact with the wall towards the back of the robot to lose their adhesion with the wall prematurely. Thus, this robot is only able to support its own mass of \qty{0.5}{kg}; it was not shown to be capable of towing additional payloads.
    Our research proposes a way to overcome the suction cup force trade-off by utilizing specially designed bistable passive suction cups to create a high-payload wall-climbing robot.

    \revcomment{}{Bistability is a proven design strategy which enables multiple functional states through a single structure, allowing systems to switch between stable configurations without continuous energy input. A common example is a light switch that exploits two stable equilibrium positions for robust, repeatable performance. Our work adapts the features of bistability to suction cups.}

    Unlike traditional passive suction cups, bistable suction cups have two stable configurations: ``flipped-up'' and ``flipped-down'' (see Fig.~\ref{fig:bistability}).
    By switching from the flipped-up to flipped-down configuration, such suction cups can generate a pressure difference without needing to be pushed into a surface.
    Bistable suction cups thus offer a solution to the suction cup force trade-off as they enable the generation of large adhesion forces without requiring large application forces that would risk pushing a wall-climbing robot off of the wall. 
    
    
    This work proves the efficacy of bistable suction cups for creating high performance wall-climbing robots.
    The primary contribution is the design and experimental validation of a track-based system that automatically engages and disengages bistable suction cups to achieve locomotion on smooth surfaces.
    The robot has limited functionality and can only move in straight lines, but it effectively demonstrates the ability to achieve vertical wall climbing on a variety of surfaces including glass, wood, metal, and painted surfaces, sideways and upside-down climbing, and is able to tow a payload of \qty{7.940}{kg} in addition to the \qty{3.532}{kg} mass of the robot itself  \revcomment{}{for a payload-to-weight ratio of 2.25} (see in Fig. \ref{fig:collage}).
    Notably, this performance is achieved using passive suction cups rather than any of the more energetically demanding active adhesion methods and achieves a much larger payload capacity than existing passive suction cup-based robots.
    
    The remainder of this paper is organized as follows. The bistable suction cup theory and design along with the wall-climbing robot design are presented in Section II. Experiments to evaluate the performance of the suction cups and the robot are described in Section III. In Section IV we discuss the results of our experiments and Section V presents our conclusions.

\section{Methods}
\subsection{Bistable suction cups}
\subsubsection{Theory}

\begin{figure}
    \centering
    \includegraphics[width=\linewidth]{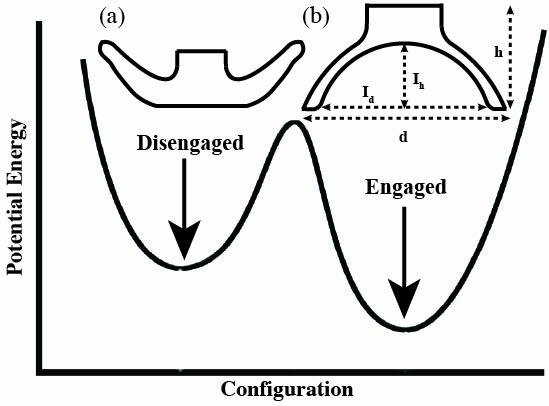} 
    \caption{Bistable suction cup design requires work to be done on the system to change configurations. (a) Disengaged position where cup is ``inverted". (b) Engaged position where cup is in the standard position. The force required to flip from disengaged to engaged is greater than in the other direction.}
    \label{fig:bistability}
\end{figure}

For a mechanism to be considered bistable, it must have two stable equilibrium positions. The transition period between the two stable equilibria is often referred to as the ``snap-through" \cite{opdahl1998investigation}. In Fig.~\ref{fig:bistability}, the potential energy curve represents how the mechanism interacts with inputs of energy. The two valleys represent the stable positions with an input of energy required to snap-through or move the bistable mechanism from one position to the other. \revcomment{}{Suction cups generate an adhesion force through a pressure difference between the air trapped underneath and the surrounding air. While a standard suction cup requires air to be forced out from under it to generate this difference, a bistable cup is able to trap very little air underneath during placement and does not require air to be pushed out to generate a pressure difference.}
Previous studies have explored the application of bistable properties in active suction cups\cite{kirsch2018bistable}. The work presents an SMA actuated bistable vacuum suction cup. There has also been work in the use of bistability in passive suction cups used in flying robots \cite{hsiao2022novel}. This work shows how bistability can be leveraged to passively attach to a smooth surface. In their application, the quad copter's weight landing triggers the bistable mechanism which creates the suction. We have adapted a similar methodology to a wall-climbing robot to overcome the limitations of traditional passive suction cups. Instead of using the weight of a quadcopter to trigger the mechanism, we use a custom track-based mechanism to physically push the cup between its stable configurations. In their robot, the bistable mechanism is separate from the suction cup; however, we incorporated bistable behavior directly into our suction cup design.

To design our bistable suction cup, we began by creating a mold for a standard suction cup. \revcomment{}{We then adjusted the radius to find the concavity that would yield a suction cup with the desired bistable behavior. Through testing, we discovered a ``cutoff" ratio, the interior diameter divided by the interior radius, between a cup being standard and bistable. 
After testing over 10 different designs with varying parameters, we landed on a ratio of $\frac{I_d}{I_h} = 2.6$}. Our design is shown in Fig. \ref{fig:bistability} with parameters outlined in Table \ref{tab:suction-cup-specs}.

To fabricate the bistable suction cup, we designed a mold using computer-aided design software; a detailed outline of our design is available at: (\url{https://github.com/free-laboratory/Wall-Climbing-Robot}). We then molded the suction cup from Smooth-On Smooth-Sil\texttrademark{} 945 \cite{hernandez2024stickiness}.
To remove air bubbles, we placed the mold in a vacuum chamber for 10 minutes before the 24 hour curing process.
The resulting bistable suction cup has two stable configurations called ``engaged" and ``disengaged" seen in the Fig.~\ref{fig:bistability}, as desired. 
\begin{table}[H]
\renewcommand{\arraystretch}{1.3}
\caption{SPECIFICATIONS OF BISTABLE SUCTION CUP}
\label{tab:suction-cup-specs}
\centering
\begin{tabular}{|c|c|c|}
\hline
 & \bfseries Engaged Position & \bfseries Disengaged Position\\
\hline
Height (mm) & 36 & 22.1\\
\hline
Diameter (mm) & 69 & 71.6\\
\hline
Interior Radius (mm) & \multicolumn{2}{c|}{23.2}\\
\hline
Interior Diameter (mm) & \multicolumn{2}{c|}{60} \\
\hline
Wall Thickness (mm) & \multicolumn{2}{c|}{4}\\
\hline
Mass (g) & \multicolumn{2}{c|}{34}\\
\hline
\end{tabular}
\end{table}

A bistable suction cup can be applied to a surface in two ways. The first one is the traditional method that is used to apply standard suction cups: an applied normal force to create the suction. The second method is to exploit the transition between the bistable positions. With the cup beginning in the disengaged position, a small force can flip the brim of the cup down and induce a snap-through transition to the engaged position.
When this occurs, the air under the cup is pushed out to the sides, leaving very little air trapped in the cavity between the surface and the cup. When a load is exerted on the suction cup, (external payload such as the robot weight, or the cup's restoring force), it draws a slight vacuum, the pressure difference creates an adhesion force. The robot uses this method to adhere the cups to the wall, requiring mechanisms to flip the cup between the configurations.

\subsection{Wall-Climbing Robot Design}
\begin{figure}
    \centering
    \includegraphics[width=0.5\textwidth]{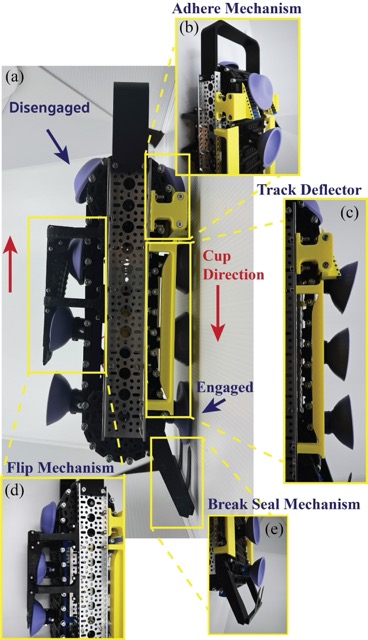} 
    \caption{(a) Full Wall-Climbing Robot Design (b) Adhere mechanism to take suction cup from disengaged to engaged position and adhere to surface (c) Track deflector to force the tread towards the robot frame (d) Flip mechanism to flip the cup from the engaged to disengaged position. (e) Break seal mechanism to detach cup from wall. Cups rotate in clockwise direction, marked in red}
    \label{fig:robot-design}
\end{figure}

To achieve wall-climbing with bistable suction cups, we designed a tread system with cups attached to it that does the following to each cup as it completes one revolution of the tread (see Fig. \ref{fig:robot-design}): (1) Switch the suction cup from the disengaged to engaged position as it comes in contact with a surface (see Fig. \ref{fig:robot-design} (b)); (2) Break the seal between the suction cup and the surface (see Fig. \ref{fig:robot-design} (e)); (3) Flip the cup from the engaged to the disengaged position to reset it (see Fig. \ref{fig:robot-design} (d)). The designs of each component of the climbing robot are described below. In our design, three suction cups are adhered to the surface at any given time. 
This provides multi-cup contact for redundancy without drastically expanding the robot's footprint and mass, balancing the trade-off between these two characteristics.

\subsubsection{Robot Frame and Tread System}
The frame, tread system, and motor for the robot are sourced from goBILDA components (www.gobilda.com). The frame is composed of an aluminum U-channel (Part \#: 1120-0012-0336) which is connected by standoffs to two parallel pieces of aluminum U-channel (Part \#: 1121-0012-0312). The motor (Part \#: 5203-2402-0188) is mounted in the central U-channel, and transmits power to the front axle using bevel gears. The tread system (Part \#: 2400-0112-0002) wraps around the frame and is supported by track sprockets (Part \#: 2401-0014-0012) mounted at the front and back of the robot. The robot is powered by an external \qty{12}{V} power supply that is connected directly to the motor.

\begin{table}
\renewcommand{\arraystretch}{1.3}
\caption{SPECIFICATIONS OF ROBOT}
\label{tab:robot-specs}
\centering
\begin{tabular}{|c|c|}
\hline
Size (mm) & 524 x 200 x 161\\
\hline
Mass (kg) & 3.532\\
\hline
Climbing Speed (cm/s) & 11.5\\
\hline
Nominal Voltage of DC Motor (V) & 12\\
\hline
Nominal Current of DC Motor (A) & 0.25\\
\hline
Nominal Torque of DC Motor (Nm) & 24.51\\
\hline
\end{tabular}
\end{table}

\subsubsection{Flip Mechanism}
To facilitate the application of the suction cup onto a surface, it is imperative for the cup to initially be in the disengaged position. The flip mechanism shown in Fig. \ref{fig:robot-design} (d) transitions the cup from the engaged position to the disengaged position. The mechanism consists of two parallel outer ramps that straddle the sides of the suction cup as they pass by. A third ramp is placed in the middle to prevent the cup from folding on itself. As the cup moves along, the edges are gradually pushed out until the cup pops to the disengaged position. The design features rollers on the ramps that minimize friction against the cup brim.
\subsubsection{Adhere Mechanism}
To adhere the cup to the surface, the mechanism utilizes two ramps, one on each tangential side of the cup, that push the edges of the cup, flipping them onto the surface. The mechanism features rollers, similar to those in the flip mechanism, to reduce friction between the cup and the ramp (see Fig. \ref{fig:robot-design} (b)). In a single motion, the cup transitions from a disengaged to an engaged position, and adheres to the surface.

A second part to assist this process, seen in yellow in (Fig. \ref{fig:robot-design} (c)), deforms the shape of the tread so that cups already adhered to the surface are pulled closer to the main structure of the robot. This in turn moves the next disengaged cup waiting to be flipped closer to the surface, which reduces the application distance between the cups and the surface. This can greatly improve the cups' ability to adhere to the surface. And as seen in Fig. \ref{fig:application-distance-graph} the closer the cup is to the surface at application, the higher the suction force.
\subsubsection{Break Seal Mechanism}
A mechanism to remove the cup from the surface is needed due to the high suction force generated by the cup. The mechanism features two  flexible sticks as seen in Fig. \ref{fig:robot-design} (e) which ride along the surface and break the seal to remove the pressure difference, thus releasing the cup from the surface.

\section{Experiments}   \label{sec:experiments}
Experiments were conducted to characterize the
bistable suction cup and evaluate the wall-climbing robot's performance. All tests were conducted at room temperature ($\sim 22\,^{\circ}\mathrm{C}$) and ideal indoor humidity ($\sim 40\%$).  The following experiments were performed: a suction cup flip test, a suction cup pull-off force test, a suction cup pull-off force test with surface roughness, a suction cup shear force sliding test with surface roughness, a suction cup hang time test, payload capacity, robot hang-time test, and a robot climbing test. This section presents the findings of each experiment.
\subsection{Suction Cup Flip Tests}
Since the suction cup used for wall-climbing was bistable, we measured the force required to flip the cup between the stable configurations. We measure the force required to flip the cup from engaged to disengaged and disengaged to engaged positions using a tensile testing machine.
\begin{figure}
    \centering
    \includegraphics[width=\linewidth]{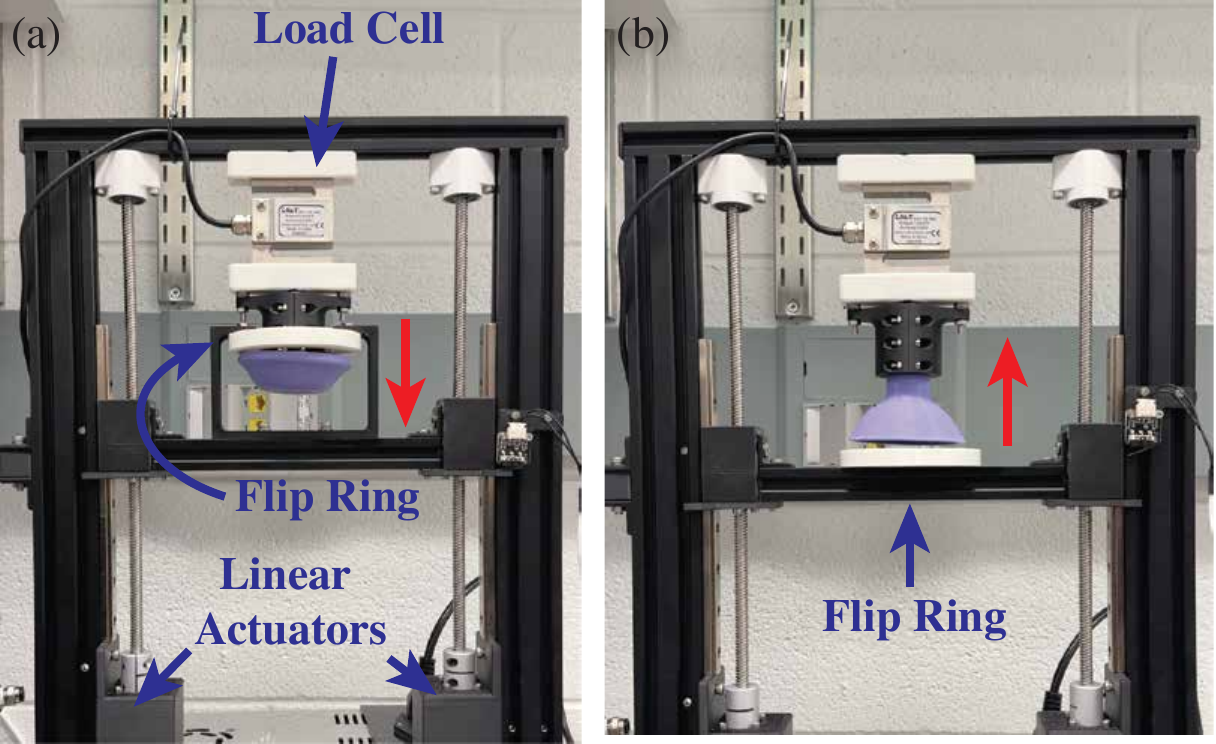} 
    \caption{The forces required to engage and disengage the suction cup were measured (a) The flip ring pulls the cup down to flip from disengaged to engaged (b) The flip ring pushes the cup up from engaged to disengaged. The machine moves flip ring in red direction.}
    \label{fig:flip-tests}
\end{figure}

\subsubsection{Disengaged to Engaged}
The cup was mounted to a single axis load cell (CALT: DYLY-103). 
First, the load cell was tared, and then the flip ring moved down towards the cup to flip it (see Fig. \ref{fig:flip-tests} (a)). The force was measured for the entire duration of the travel until the cup is flipped. The force required to flip the cup was averaged over five trials, in which the cup was reset by hand between each trial.

\subsubsection{Engaged to Disengaged}
The reverse was done for the engaged to disengaged position, similar to Part 1 of the experiment. The flip ring was mounted below the cup (see Fig. \ref{fig:flip-tests} (b)) and moved upward to flip.

The force required to flip the suction cup from the engaged position to the disengaged position was measured to be $36.23 \pm 0.4\, \mathrm{N}$, which is approximately 4x the force required to flip the suction cup from the disengaged position to the engaged position $9.71 \pm 0.8\, \mathrm{N}$. This is due to the cross-sectional shape of the suction cup design, parameters could be altered to reduce the respective force required to flip the cup between the stable configurations.

\subsection{Suction Cup Pull-off Force Test}
\begin{figure}
    \centering
    \includegraphics[width=0.5\textwidth]{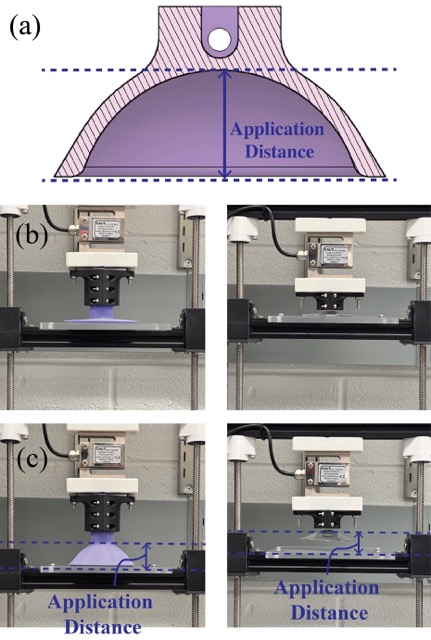} 
    \caption{The pull off forces at different application distances were tested on two types of suction cups. Bistable (purple) suction up on the left and standard clear PVC suction cup on the right (a) The Application Distance measures the distance between the surface and the top of the interior dome of the suction cup (b) \qty{0}{mm} Application Distance (full compression) (c) \qty{17}{mm} Application Distance (little to no compression)}
    \label{fig:application-distance}
\end{figure}

Previous work has tested how the distance a cup is deformed affects the detachment force \cite{Kawasaki2014DevelopmentOA}.
We tested the pull-off force for our bistable suction cup at various applied distances with respect to the surface defined in Fig \ref{fig:application-distance} (a).
We defined 0 as the position where our suction cup is fully pressed into the surface such that its center is touching (see Fig. \ref{fig:application-distance} (b)), and the maximum distance to be the distance at which the brim of the cup first makes contact with a surface (see Fig. \ref{fig:application-distance} (c)). We then tested the distance from 0 to \qty{17}{mm} for our suction cup. The same was done for a standard suction cup (Sankoly: SK2401SHSCR5565) which has the same diameter in its uncompressed state, but the testable range for that cup was only \qty{0}{mm} to \qty{6}{mm}. The experiment consisted of 5 trials at \qty{1}{mm} intervals of the application distance.

\begin{figure}
    \centering
    \includegraphics[width=\linewidth]{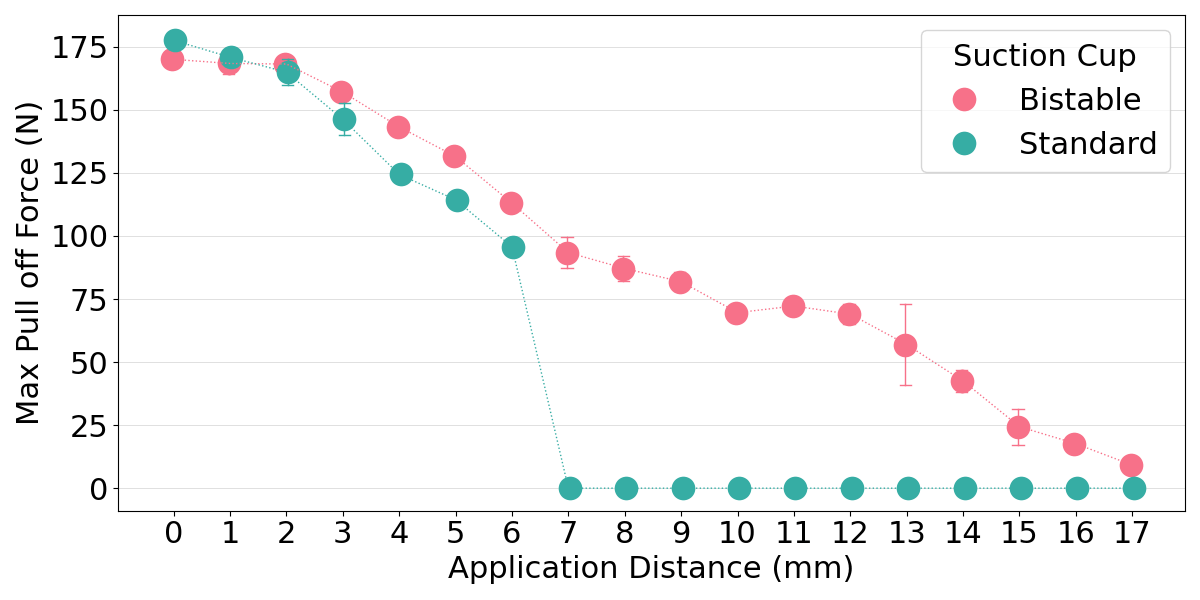} 
    \caption{Maximum pull off force at various application distance averaged over 5 trials for bistable (pink) and standard (aqua) suction cups were measured. Beyond \qty{6}{mm} the forces for the standard PVC cup fall to zero because it no longer makes contact with the surface due to its shallower design.}
    \label{fig:application-distance-graph}
\end{figure}
Fig. \ref{fig:application-distance-graph} shows the pull off force varying as the application distance varies. For both the bistable and standard suction cups, the lower the application distance, the greater the pull off force. Note that for \qty{0}{mm} application distance the standard suction cup had an applied force of \qty{21.4}{N} to achieve suction. Our bistable suction cup had \qty{0}{N} applied force, but requires \qty{9.71}{N} to flip the cup from disengaged to engaged where suction is achieved, as measured in the flip tests in Section~\ref{sec:experiments}-A.

\subsection{Normal Force Surface Roughness}
We tested our bistable suction cup on a sweep of sandpaper grits to get an understanding of how its performance varies as a function of surface roughness. Using an application distance of \qty{0}{mm}, the cup was adhered to an acrylic platform covered in sandpaper of various grits. The platform moved down away from the suction cup at a constant rate of \qty{1}{mm/second} until losing contact with the cup. Trials were repeated five times on 600, 800, 1000, 1500, 2500, 3000, 5000, 7000 grit sandpaper, and acrylic. For lower grit sandpaper (corresponding to the highest surface roughness) the leak rate of the suction cup limited the maximum pull force we could measure before suction was completely lost. The leak rate of the PVC cup, used in the previous experiment, was deemed too fast to have any measurable data.
\begin{figure}
    \centering
    \includegraphics[width=\linewidth]{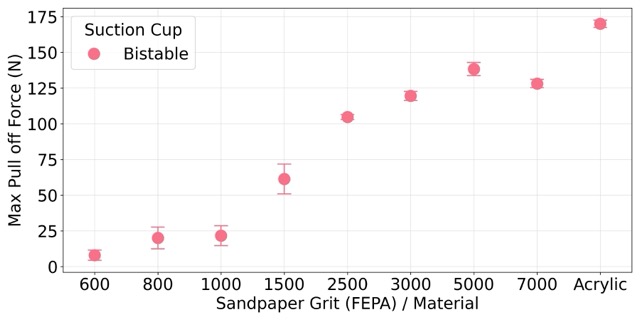} 
    \caption{Maximum pull off force at various sandpaper grits averaged over 5 trials for bistable (pink) suction cup were measured.}
    \label{fig:normal-force-sandpaper}
\end{figure}

\begin{figure}
    \centering
    \includegraphics[width=\linewidth]{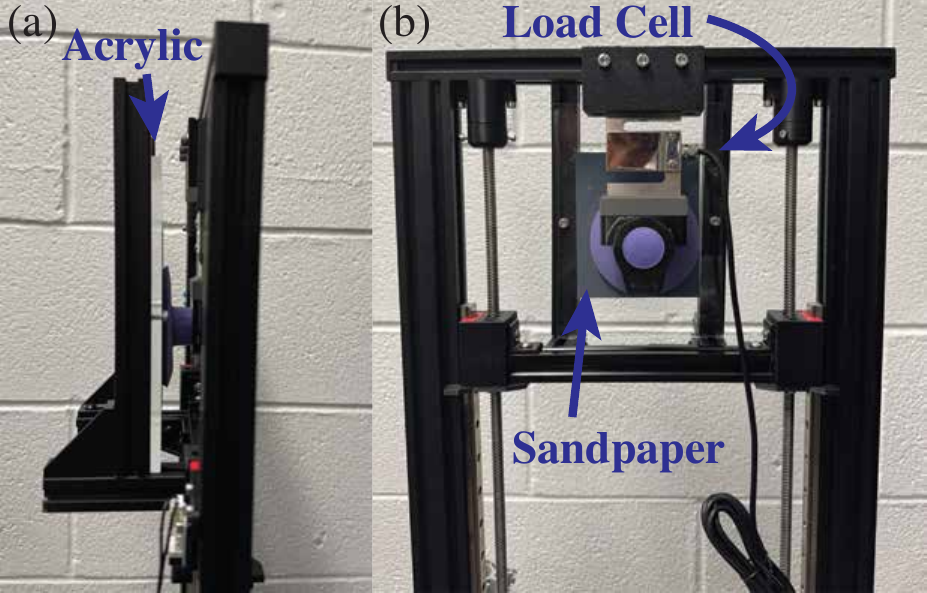} 
    \caption{The sliding force on different surface roughness levels was tested on the bistable suction cup. (a) The side view of the cup adhered to the test setup. (b) The back view of the cup adhered to the sandpaper.}
    \label{fig:shear}
\end{figure}
\subsection{Shear Force Surface Roughness}
\begin{figure}
    \centering
    \includegraphics[width=\linewidth]{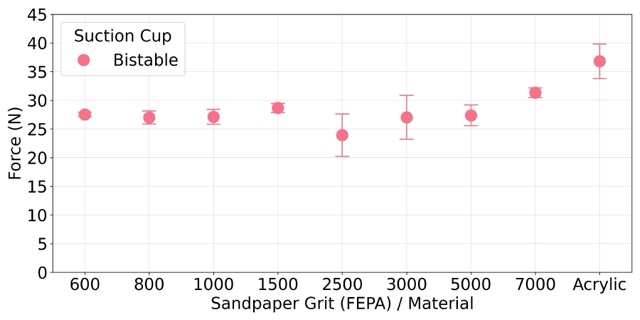} 
    \caption{Force required to begin sliding at various sandpaper grits and acrylic averaged over 5 trials for bistable (pink) suction cup were measured.}
    \label{fig:shear-force}
\end{figure}

Fig. \ref{fig:shear} shows the setup for measuring the force required to cause the cup to begin sliding on the test surface in the shear direction. Fig.~\ref{fig:shear-force} shows how the suction cup performs on a variety of sandpaper grits as well as acrylic. Trials were repeated five times. The values proved to be fairly consistent between the different sandpaper grits and the acrylic.

\subsection{Suction Cup Hang Tests}
\begin{figure}
    \centering
    \includegraphics[width=\linewidth]{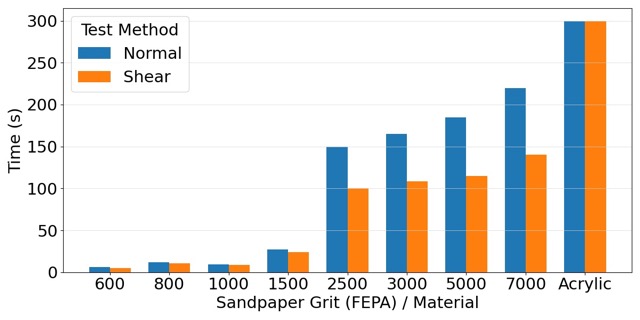} 
    \caption{Average fall-off time of the bistable suction cup with a \qty{1}{kg} hanging mass across different sandpaper grits and acrylic, based on three trials blue shows the cup in shear conditions, orange shows the cup in normal conditions.}
    \label{fig:cup-hang-time}
\end{figure}

Experiments were done to measure the time a suction cup could stay adhered to a surface for both normal and shear conditions. To start the trial, a cup with a \qty{1}{kg} mass hanging from it was flipped from the disengaged to the engaged position to adhere it to the test surface. The hang time was measured in seconds and capped at \qty{300}{seconds}. Trials were repeated three times. Fig. \ref{fig:cup-hang-time} shows the results of the experiment. 

\subsection{Payload Capacity Test}
A key focus of the robot design was to test the payload capacity while climbing a vertical surface using the bistable suction cups (see Fig. \ref{fig:collage} (d)). The base robot weighed \qty{3.532}{kg} and a chain was connected to the robot to allow for additional mass to be attached. Mass was added until the robot could not climb with the added mass. The robot was reliably able to carry an additional \qty{7.940}{kg} which is a payload-to-weight ratio of 2.25 for a total of \qty{11.472}{kg}. When a heavier payload of \qty{9.071}{kg} was applied, for a total of \qty{12.613}{kg}, the robot was unable to climb.

\subsection{Robot Hang Time Test}
\begin{figure}[t]
    \centering
    \includegraphics[width=0.9\linewidth]{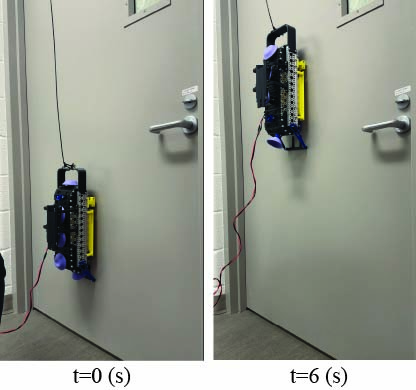}
    \caption{A wall-climbing experiment was done on a steel door, left shows the starting position and right shows the final position. The black cable attached to the top of the robot was a safety cable to prevent it from falling when detached from the door and does not have any tension or provide any structural support to the robot.}
    \label{fig:robot-climbing-door}
\end{figure}

A test was done to see how long the robot could stay stationary on a wall before falling off. This test indirectly measures the leaking rate of the cup design in a real-world climbing scenario, thus providing an informative performance benchmark. In a real-world application the robot may need to remain adhered to the wall while stationary for a significant period of time to accomplish tasks. The robot climbed up a steel door under its own power (see Fig. \ref{fig:robot-climbing-door}), then stopped and remained adhered to the wall for 3 hours and 45 minutes before falling off (no payload).

\subsection{Robot Wall-Climbing Demonstrations}
The robot successfully demonstrated wall-climbing on a variety of surfaces, including sideways and upside-down climbing.
To start, the robot is positioned by a human on the climbing surface. Once positioned, the cups can be manually engaged so the robot adheres to the surface and is ready to climb. Vertical climbing tests were done on a steel door, painted wall, wooden door, and glass window (see Fig. \ref{fig:collage}). The robot was also tested on the underside of a lab bench where the robot hung from the bench as it translated. The robot was tested traversing a wall sideways on a whiteboard. A final demonstration was done with an eraser mounted to the back of the robot which erased text written on the board as the robot moved past. This demonstration shows the robot is able to apply a force into the surface while continuing to move on the surface.
Footage of all these demonstrations can be found in the supplementary video file.

\section{Discussion}
\subsection{Explanation of experimental results}
The suction cup performance steadily decreases as the sandpaper grit decreases (surface roughness increases). This is to be expected as suction cups are based on creating a seal with the surface, and the more imperfections the surface has, the worse the seal generated will be. We see the phenomenon in Fig. \ref{fig:normal-force-sandpaper} and Fig. \ref{fig:cup-hang-time}. Both the max pull off force and the time before detaching decrease with more rough surfaces. 
The data indicates, however, that the robot could likely climb a wall lined with 2500 grit sandpaper as the shear force it generates would be sufficient to support the robot's body-weight, and the hang-time is sufficiently long to maintain adhesion throughout a complete revolution of the tread system.

Sandpaper was used as a way to test surface roughness by taping the sandpaper to an acrylic piece. We were unable to find an exact relationship between the grits of sandpaper we used and surface roughness $R_a$ \cite{gadelmawla2002roughness}. However some rougher sandpapers have experimentally verified $R_a$ values (\url{https://www.ljstar.com/resources/surface-finish-charts/}).

In the hang-time test, the robot remained on the steel door seen in Fig. \ref{fig:robot-climbing-door} for 3 hours and 45 minutes.
While this experiment demonstrates long-term passive adhesion on relatively smooth surfaces, this time would vary depending on how well the suction cup is sealed against the surface. The time would change based on the surface interacting with the suction cup. Varying surface conditions affects the leak rate of the suction cup as seen in Fig. \ref{fig:cup-hang-time} which could affect the time. Due to the design and the nature of the bistable suction cup, its nominal volume is much larger than standard cups, which allows for more air flowing back into the suction cup before it detaches. The hang time also depends on the mass of the robot and its payload. A higher force pulling on the cup would likely increase the leak rate of air flowing into the suction cup, thus affecting the resulting hang time.

\subsection{Current limitations}
Despite the improvements in wall-climbing using passive suction cups described in this paper, there are current limitations of the system. As with most suction cup designs, the robot only works on relatively smooth surfaces. Surfaces such as brick or porous concrete would be  impossible for the robot to climb. 
Since the suction cups rely on creating a seal with the surface, if there are any gaps or valleys in the surface that are beyond the conformability of the suction cup, it will not be able to create any suction. 

The flipping and un-flipping mechanism for the bistable suction cups only allows the robot to move in one direction, which could lead to potential issues in a real-world setting because it cannot return to the starting position. A related limitation is the inability of the robot to turn.
The current design moves in a singular straight line direction. Achieving turning would greatly improve the maneuverability of the robot allowing it to reach more areas when climbing.

Currently, the robot requires being propped against the wall by a human before it can climb on its own. While the design proves the concept of using bistable suction cups to climb a wall, in the real world it would be useful for the robot to be able to start from the ground and move onto the wall autonomously, or be able to transition from wall-to-wall and wall-to-ceiling without human intervention.

\subsection{Future Improvements and Next Steps}
In the future, improvements could be done to optimize the design of the bistable suction cup for improved adhesion. Some minor adjustments in the parameters of the robot design could further optimize the payload capacity of the robot. Although achieving a high payload is a priority, minimizing the structural mass of the base robot would likely yield a higher capacity for external loads. A battery could be mounted to the robot to provide power to the motor which would make it untethered. Additional work could be done to make the cups more robust to surface imperfections. More substantial alterations could be done to the robot to make it capable of turning. 

\section{Conclusion}
In conclusion, this work presents a novel method for wall-climbing robots that leverages bistable suction cups to overcome the challenges faced by traditional passive adhesion methods. By utilizing bistable passive suction cups, which require no pushing force to initiate adhesion, the robot achieves high-payload capacity and energy efficiency, addressing the limitations of existing designs. The robot's ability to climb on smooth surfaces, such as glass, wood, metal, and painted surfaces, its capacity to carry a \qty{7.940}{kg} payload (for a payload-to-weight ratio of 2.25), and its ability to climb inverted and sideways, demonstrates the effectiveness of bistable suction cups in enhancing the performance of wall-climbing robots. This advancement constitutes a significant step towards automating labor-intensive tasks like inspection and maintenance on vertical surfaces due to its efficiency and reliability in real-world applications.


\bibliographystyle{IEEEtran}
\bibliography{references}

\end{document}